\pdfoutput=1
\documentclass[]{spie}  

\usepackage{amsmath,amsfonts,amssymb}
\usepackage{graphicx}
\usepackage{subfig}
\usepackage[table]{xcolor}
\usepackage[colorlinks=true, allcolors=blue]{hyperref}
\usepackage{lscape}
\usepackage{booktabs}
\usepackage{tabularx}
\usepackage{todonotes}

\usepackage{orcidlink}

\title{Multi-Objective Deep-Learning-based Biomechanical Deformable Image Registration with MOREA}

\author[a]{Georgios Andreadis\,\orcidlink{0000-0002-8955-5939}}
\author[b]{Eduard Ruiz Munné\,\orcidlink{0000-0002-6061-8609}}
\author[b]{Thomas H. W. Bäck\,\orcidlink{0000-0001-6768-1478}}
\author[c]{\\Peter A. N. Bosman\,\orcidlink{0000-0002-4186-6666}}
\author[a]{Tanja Alderliesten\,\orcidlink{0000-0003-4261-7511}}
\affil[a]{Dept. of Radiation Oncology, Leiden University Medical Center (LUMC), The Netherlands}
\affil[b]{Leiden Institute of Advanced Computer Science (LIACS), Leiden University, The Netherlands}
\affil[c]{Evolutionary Intelligence Group, Centrum Wiskunde \& Informatica (CWI), The Netherlands}

\authorinfo{Correspondence: Georgios Andreadis (G.Andreadis@lumc.nl) and/or Tanja Alderliesten (T.Alderliesten@lumc.nl).}

\begin{document} 
\maketitle

\begin{abstract}
When choosing a deformable image registration (DIR) approach for images with large deformations and content mismatch, the realism of found transformations often needs to be traded off against the required runtime.
DIR approaches using deep learning (DL) techniques have shown remarkable promise in instantly predicting a transformation.
However, on difficult registration problems, the realism of these transformations can fall short.
DIR approaches using biomechanical, finite element modeling (FEM) techniques can find more realistic transformations, but tend to require much longer runtimes.
This work proposes the first hybrid approach to combine them, with the aim of getting the best of both worlds.
This hybrid approach, called DL-MOREA, combines a recently introduced multi-objective DL-based DIR approach which leverages the VoxelMorph framework, called DL-MODIR, with MOREA, an evolutionary algorithm-based, multi-objective DIR approach in which a FEM-like biomechanical mesh transformation model is used.
In our proposed hybrid approach, the DL results are used to smartly initialize MOREA, with the aim of more efficiently optimizing its mesh transformation model.
We empirically compare DL-MOREA against its components, DL-MODIR and MOREA, on CT scan pairs capturing large bladder filling differences of 15 cervical cancer patients.
While MOREA requires a median runtime of 45 minutes, DL-MOREA can already find high-quality transformations after 5 minutes.
Compared to the DL-MODIR transformations, the transformations found by DL-MOREA exhibit far less folding and improve or preserve the bladder contour distance error.
\end{abstract}



\section{INTRODUCTION}

Many applications in image-guided radiation treatment could benefit from the transfer of information between multiple medical images of the same patient. 
However, local tissue deformations or content mismatch can hinder this transfer, requiring deformable image registration (DIR) to establish a spatial correspondence between the images.
In this work, DIR entails registering a source image onto a target image.
As this task has been shown to be inherently multi-objective~\cite{Pirpinia2017}, a set of conflicting objectives should be optimized, simultaneously.
The aim of multi-objective DIR is to find a so-called approximation set of transformations that covers the trade-off front of Pareto optimal transformations spanned by these objectives.
To be of value, the transformations must be biomechanically plausible (i.e., realistic).
Furthermore, to be useful in various clinical applications, they typically need to be calculated within a short time frame.
These two requirements, however, can be in conflict.

Two predominant types of DIR approaches have emerged, each excelling in one of these two requirements.
On the one hand, DIR approaches using deep learning~(DL) techniques~\cite{Eppenhof2020,Salehi2022,DeVos2019}, such as the recently introduced multi-objective DL-MODIR approach~\cite{Grewal2024}, can instantly predict a transformation for a given registration problem, after being trained on a set of prior problems.
On problems with large deformations and content mismatch, however, this speed often comes at the expense of transformation realism, as current prediction models are not grounded in a realistic transformation model.
On the other hand, DIR approaches which use biomechanical, finite element modeling~(FEM) techniques~\cite{Brock2005,Zhong2012,Zhang2004} can find more realistic transformations, but often require a longer runtime.
One example is the recently proposed MOREA~\cite{Andreadis2023}, an evolutionary algorithm-based, multi-objective DIR approach using a FEM-like biomechanical mesh transformation model.
Until now, these two types of DIR approaches have always been considered separately.
On the path to widespread clinical adoption of DIR, however, an approach is needed that overcomes the limitations of both approach types.

\begin{figure}[t]
    \centering
    \includegraphics[width=\linewidth]{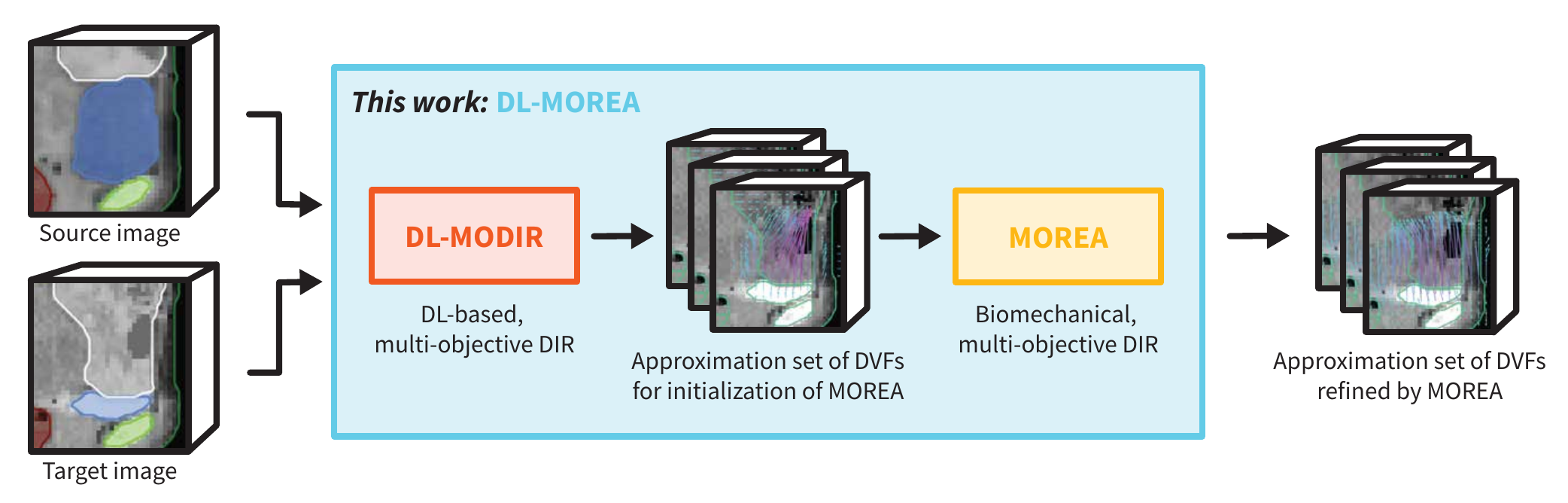}
    \caption{Illustration of the proposed hybrid DL-MOREA DIR approach and its main components.}
    \label{fig:approach-illustration}
\end{figure}

In this work, we introduce a hybrid approach to combine the best of both worlds.
The hybrid, called \mbox{DL-MOREA} and illustrated in Figure~\ref{fig:approach-illustration}, uses the transformations found by the DL-based DL-MODIR approach to smartly initialize the biomechanical FEM-like mesh model used by the MOREA approach, to more efficiently optimize the model.
The MOREA approach then refines these transformations, reducing the required runtime for MOREA to reach high-quality solutions.
Because of the multi-objective nature of both approaches, a comprehensive transfer of information is possible.
We compare the hybrid to its components in experiments on pelvic CT scans of 15 patients by analyzing their transformations quantitatively and qualitatively.

\section{MATERIALS AND METHODS}

\begin{figure}[b]
    \centering
    \includegraphics[width=\linewidth]{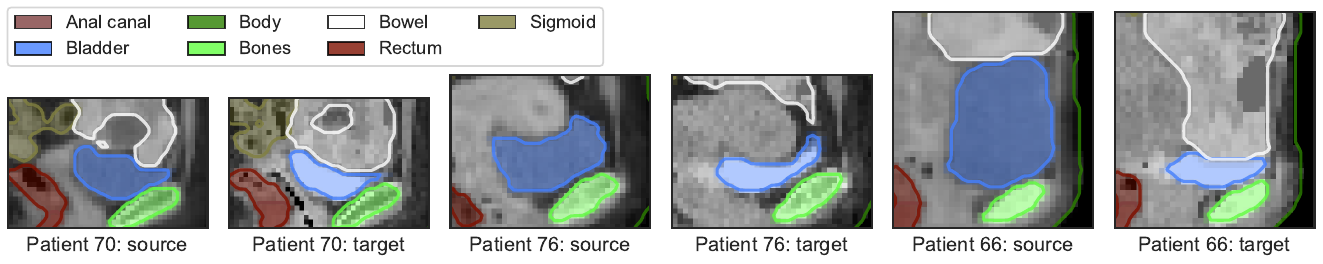}
    \caption{\footnotesize Reference source and target renders of the three selected patients. Shown are annotated sagittal slices.}
    \label{fig:patient-reference-renders}
\end{figure}

\subsection{Dataset}
The registration problems in this study are derived from pelvic CT scans of 75 cervical cancer patients, acquired for radiation treatment planning purposes, using a Philips Brilliance Big Bore scanner.
For each patient, we include a source image with a full bladder and a target image with an empty bladder, taken shortly thereafter.
For better visibility, the empty bladder was typically imaged after a bladder fluid contrast agent was administered.
This was the case in 64 of the 75 patients.
On both source and target images, 7 organs are delineated by a radiation therapy technologist~(RTT), as seen in Figure~\ref{fig:patient-reference-renders}.
The image pair is resampled to a resolution of $3 \times 3 \times 3$~\textit{mm} and rigidly registered to align the bony anatomies, using the bone contours.
Both images are then cropped to an axis-aligned bounding box around the bladder with a 30~\textit{mm} margin, using the maximal bounds from either image.
To comply with the dimensionality requirements of DL-MODIR, all images are padded to in-plane dimensions of $96 \times 96$ voxels and a minimum depth of 48 voxels.
If images exceed a depth of 48, overlapping fragments of depth 48 are prepared.
The total patient cohort is divided into a training and validation set of 60 patients, with no. 2--61, and a test set of 15 patients, with no. 62--76.

\begin{figure}[t]
    \centering
    \includegraphics[width=0.45\linewidth]{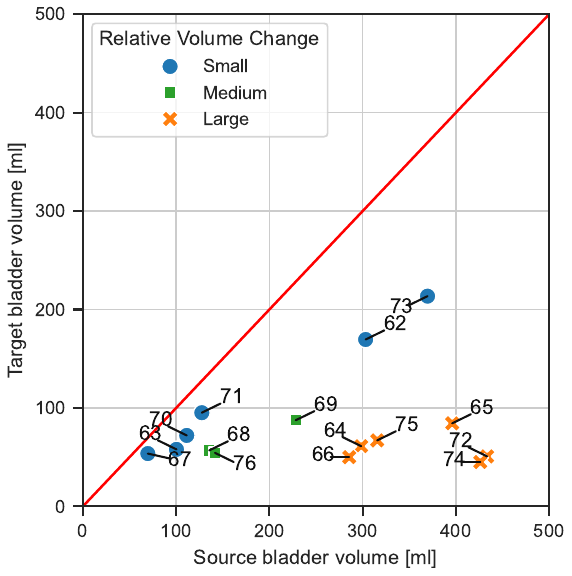}
    \caption{Scatter plot of the source and target volumes of the bladder of each tested patient. Patients are clustered into three clusters by their relative volume change between source and target, indicated in color in the plot. The red line indicates unchanged volume, while all data points below it reflect patients whose target bladder volume is smaller than their source bladder volume.}
    \label{fig:bladder-volume-scatter}
\end{figure}

From the test set, three representative patients are automatically chosen for illustration by clustering the patients on their relative bladder volume change $V_t / V_s$, where $V_s$ is the source bladder volume and $V_t$ is the target bladder volume.
A $V_t / V_s$ value of 1.0 therefore indicates no change between the bladder volumes, while a value lower than 1.0 represents a target bladder volume that is smaller than its corresponding source bladder volume.
We employ this relative change as an indicator of their registration difficulty.
Using K-Means clustering, we identify three clusters, with a small, medium, and large relative volume change, respectively.
The relative bladder volume change of each patient and their corresponding cluster is depicted in Figure~\ref{fig:bladder-volume-scatter}.
The derived clustering thresholds of the relative bladder volume change metric are: 0.5 (between the small change cluster and the medium change cluster), and 0.3 (between the medium change cluster and the large change cluster).

\subsection{DIR Approaches}

\paragraph{DL-MODIR}

This approach combines a recently proposed multi-objective learning technique~\cite{Deist2023} with the well-known DL-based VoxelMorph approach~\cite{Balakrishnan2019} to perform multi-objective, DL-based DIR.~\cite{Grewal2024}
A set of encoders and decoders are trained to predict transformations that together maximize the hypervolume~\cite{Zitzler2008} of the transformations in the approximation set.
Three loss functions are optimized simultaneously: (1) the smoothness of the transformation (the squared sum of spatial gradients of the deformation), (2) the image similarity (the normalized cross correlation between target and transformed source image), and (3) the organ segmentation similarity (the Dice score of target and transformed source organ segmentations).
As shown in Figure~\ref{fig:patient-reference-renders}, the following organ segmentations are modelled: anal canal, bladder, body, bones, bowel, rectum, and sigmoid.
The output of this approach is a set of 15 transformation deformation vector fields (DVFs), each representing a different trade-off of these loss functions.

\paragraph{MOREA}

This approach employs a biomechanical, dual-dynamic mesh transformation model in which two corresponding meshes are used, one for each image.~\cite{Andreadis2023}
These meshes are deformed through optimization with a modern, GPU-parallelized evolutionary algorithm (EA), which enables MOREA to capture large deformations and content (dis)appearances.
The EA optimizes a population of individuals, each representing a transformation.
Three objectives are considered simultaneously: (1) the deformation magnitude (the relative deformation of mesh edges, using Hooke's law), (2) the image similarity (the squared sum of intensity differences between target and transformed source images), and (3) the organ contour distances (the point-wise distances between contour point sets of target and transformed source segmentations), for the same organ contours as used by DL-MODIR.
The different material properties of organs and bony anatomy are considered by modelling their heterogeneous elasticities.
Specifically, the bladder and bowel are both considered highly elastic and assigned an elasticity constant of 0.1, the bones are considered highly rigid and assigned an elasticity constant of $10^4$, and all other tissue is assigned a uniform constant of 1.0.
These constants were determined through preliminary experiments.
After optimization, the population-based approach returns a set of mesh pairs representing different transformation trade-offs.

\paragraph{\textit{Hybrid:} DL-MOREA}

The proposed hybrid uses the DVFs found by DL-MODIR to initialize the population of transformations in MOREA, as illustrated in Figure~\ref{fig:approach-illustration}.
The population of transformations in this approach is initialized as a set of identical copies of the original mesh configuration, i.e., the mesh as it was created during mesh generation~\cite{Andreadis2023}, with its original point coordinates preserved.
Each DL-MODIR DVF is then applied to the source meshes of a proportionally sized set of individuals in the MOREA population, multiplied with linearly increasing scalars (between $[0,1]$) for diversification.
The DVFs are applied incrementally in 100 small steps, iterating over all mesh points at each step and only applying the locally interpolated deformation vector to a mesh point if it does not result in a fold.
Mesh point order is randomized at each step to reduce order bias.
The target mesh is still initialized with random noise, as in the original initialization method used in MOREA.
It is ensured that each DVF is fully applied to at least one individual without random perturbation, to prevent information loss in the transfer between registration approaches.

\subsection{Experimental Setup}
The DL-MODIR approach is trained for 150,000 iterations, with a batch size of 1 and a learning rate of 0.0001, using 5-fold cross validation. 
After training, the model of the first fold is used for further evaluation.
MOREA is configured to perform multi-resolution DIR, starting with 200 mesh points and refined once to a finer mesh by placing intermediate points on edges of the coarse starting mesh and thereby subdividing the tetrahedra of the coarse mesh~\cite{Andreadis2022}.
The number of mesh points in this configuration was determined through preliminary experiments.
Registration results obtained with the coarse mesh are used to initialize deformations of the finer resolution mesh.
In the automatically generated coarse mesh, 20\% of the point budget is allocated to bladder surface modelling, 40\% to points on other organ contours, and the rest of the budget to mesh quality enhancements, by placing points in regions with lower point density.
This distribution of the point budget was empirically determined for the anatomical site at hand.
The mesh is cropped to the field of view, thereby excluding padded empty regions required for DL-MODIR.
Optimization with MOREA is performed using 10 clusters, a population size of 600 registration solutions, and an elitist archive of 1000 registration solutions.
This optimization is performed on Dell Precision 7920R machines with NVIDIA RTX A5000 GPUs.

The MOREA and DL-MOREA approaches are repeated 5 times for each patient with different random seeds, in the interest of reproducibility.
Qualitative results are reported for the first run of each configuration, while quantitative results are reported for the median and standard deviation values of all runs.
The DL-MODIR approach is only run once as inference is deterministic.
The training procedure is not deterministic, but was not repeated due to time constraints.
During the MOREA and DL-MOREA approaches, an early stop after 5 minutes is simulated by recording the state at that time point.
Preliminary experiments indicated that this duration was sufficient for MOREA to successfully model the main components of the deformation.
We report on outcomes both after the simulated early stop time-point and after full convergence.

\subsection{Evaluation}
From each approximation set, one transformation is automatically selected for reproducibility, by sorting the transformations by their segmentation (or contour) similarity objective and choosing the transformation with the second-best value.
The cardinality of approximation sets differs strongly between approaches, with hundreds of solutions being found by (DL-)MOREA but only 15 registration solutions being found by DL-MODIR, making it difficult to choose a consistent automatic selection strategy.
The chosen strategy proved robust enough in preliminary experiments against overfitting on one objective, while still accurately capturing the exploitative strengths of each approach.
We performed both a qualitative and a quantitative comparison of the chosen transformations.

\subsubsection{Qualitative comparison}
For a qualitative comparison, the transformed contours and inverse DVFs are inspected and compared.
Transformed contours are visualized by transforming the corresponding organ mask and applying a Gaussian blur with a standard deviation of 0.6 voxels for smoothing.

\subsubsection{Quantitative comparison}
Quantitatively, we compare the Hausdorff 95th percentile distance between the target contour and the transformed source contour of the bladder.
In this comparison, we also test for statistical significance of any observed difference between MOREA and the proposed hybrid, DL-MOREA. The relative change between the medians of two distance distributions for a certain patient and time point is reported (i.e., $1-(H_{\text{MOREA}} / H_{\text{DL-MOREA}})$, with $H_x$ being the median Hausdorff 95th percentile distance for a certain approach configuration $x$).
This is tested using the Mann-Whitney U test, as we do not assume a normal distribution of measurements, using an alpha level of 0.05.
In addition to contour distance, we also compare the inverted (i.e., folded) volume of the inverse DVF, computed as the image volume for which the Jacobian of the DVF is negative.
Moreover, we show convergence plots and report required runtimes, for the optimization runs performed by MOREA and DL-MOREA.

\begin{figure}[b]
    \centering
    \includegraphics[width=\linewidth]{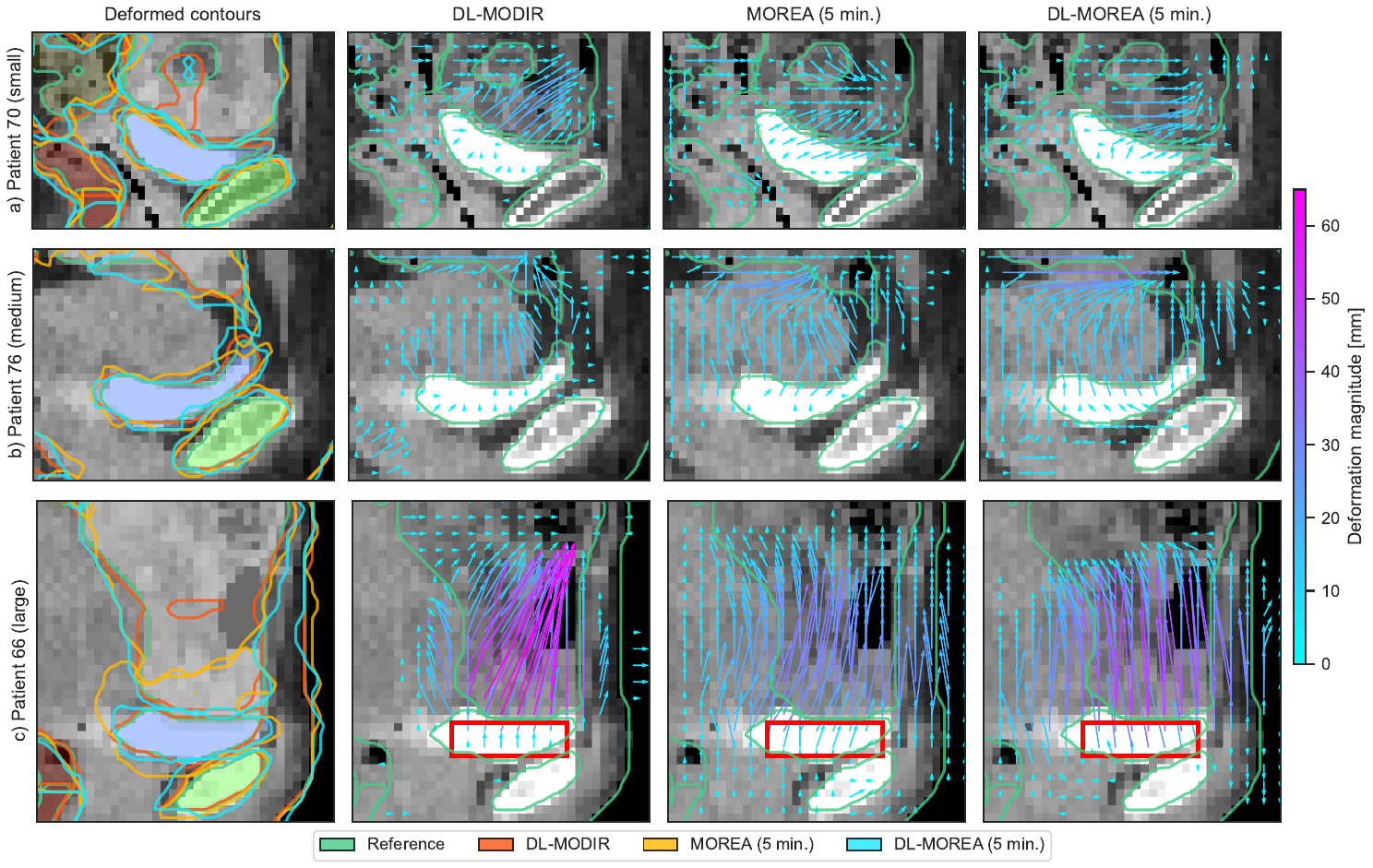}
    \caption{Renders of the found transformations of all three approaches, showing transformed contours and inverse DVFs of automatically selected registrations. For MOREA and DL-MOREA, an early stop was simulated by storing the intermediate results of optimization after 5~min. Shown are annotated sagittal slices. Arrow color indicates deformation magnitude (see legend). The red rectangle indicates a region of interest in the analysis.}
    \label{fig:patient-intermediate-solution-renders}
\end{figure}

\begin{figure}[b]
    \centering
    \includegraphics[width=\linewidth]{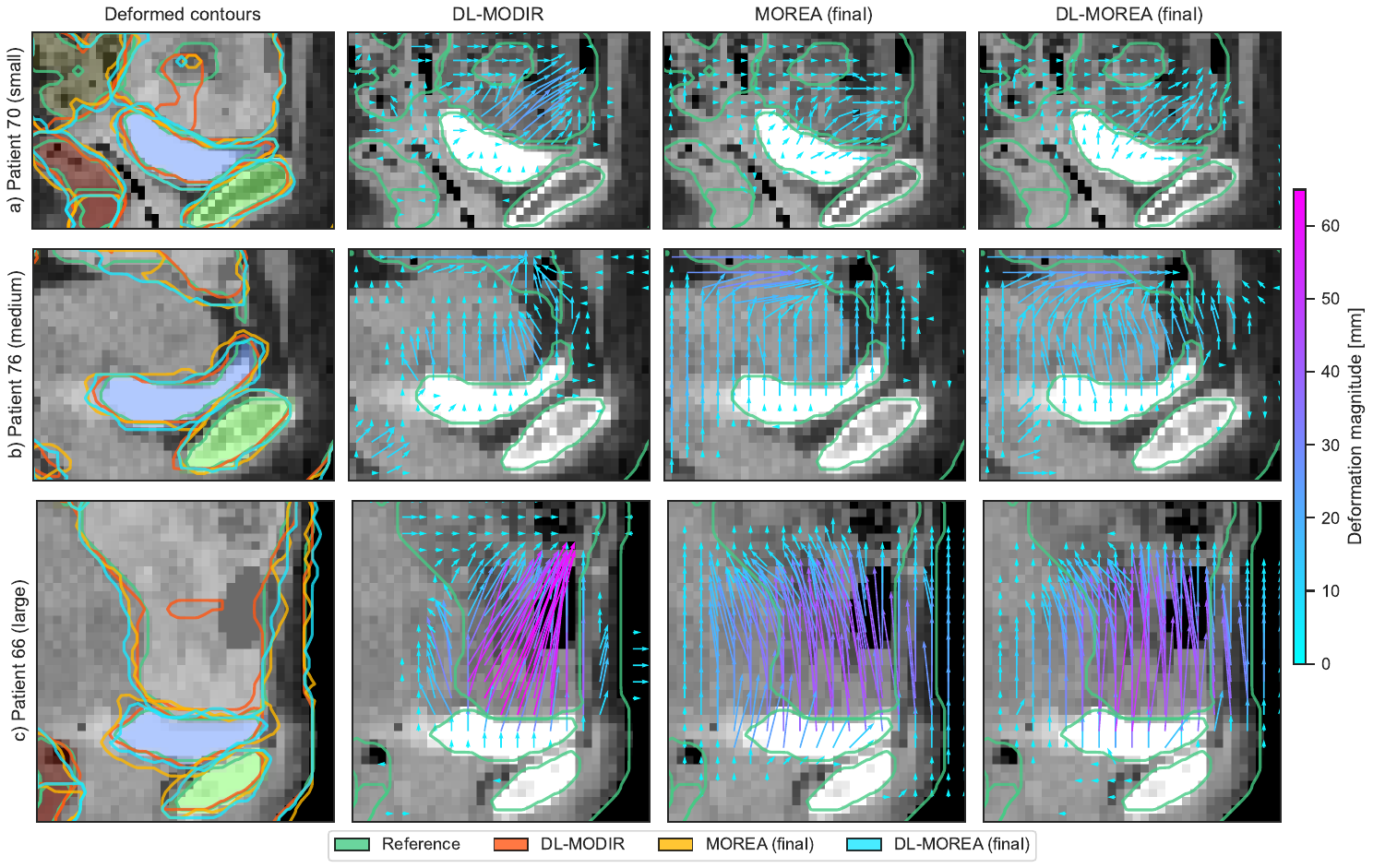}
    \caption{Renders of the found transformations of all three approaches after convergence, showing transformed contours and inverse DVFs of automatically selected registrations. Shown are annotated sagittal slices. Arrow color indicates deformation magnitude (see legend).}
    \label{fig:patient-final-solution-renders}
\end{figure}

\begin{figure}
    \centering
    \includegraphics[width=0.91\linewidth]{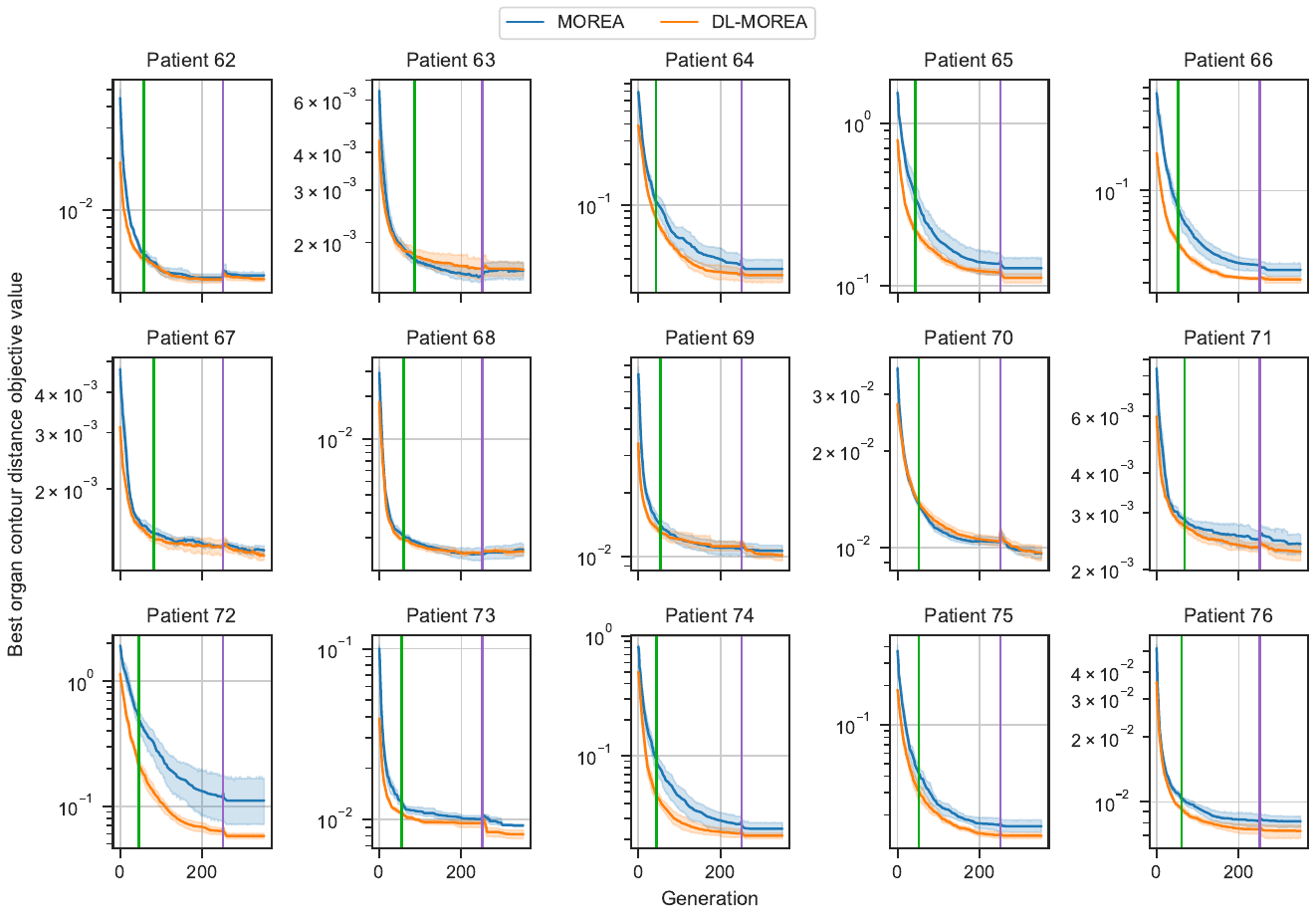}
    \caption{Convergence plots of the MOREA and DL-MOREA approaches, for all patients. The best organ contour distance objective value is plotted for each generation, aggregated across 5 repeats. The 95\% confidence interval is indicated by a shaded area. The first (green) vertical line indicates the 5-minute time point in optimization, the second (purple) vertical line indicates the resolution change from the first (coarse) resolution to the second (finer) resolution. Note that the vertical axis is scaled logarithmically.}
    \label{fig:convergence}
\end{figure}

\section{RESULTS}

\subsection{Qualitative comparison}
We show the annotated images of the three selected patients in Figure~\ref{fig:patient-reference-renders}.
In Figure~\ref{fig:patient-intermediate-solution-renders}, we show the transformed contours found by the three approaches and their underlying transformations as captured after 5~minutes, to assess the impact of the DL-based initialization.
The fully converged results are shown in Figure~\ref{fig:patient-final-solution-renders}.
Across all three patients, it can be observed that the bladder deformation is modeled in a more uniformly spread fashion by MOREA and DL-MOREA (see area marked in red in Figure~\ref{fig:patient-intermediate-solution-renders}) already after 5~minutes of optimization, while the DL-MODIR transformations are mainly facilitated by movement outside the bladder.
On the patient with a large relative difference in bladder volume depicted in row~\ref{fig:patient-intermediate-solution-renders}c, only the contours transformed by the hybrid DL-MOREA approach approximate the reference well, with MOREA not yet having converged after 5 minutes, and the transformation of DL-MODIR leaving bladder mass in the bowel region.
On patients with a smaller change (rows~\ref{fig:patient-intermediate-solution-renders}a and \ref{fig:patient-intermediate-solution-renders}b), the differences in contour quality between approaches are smaller, even though the underlying deformations still differ strongly.

When comparing the early stop results of Figure~\ref{fig:patient-intermediate-solution-renders} with the converged results of Figure~\ref{fig:patient-final-solution-renders}, small refinements such as in the transformed contours of DL-MOREA in row \ref{fig:patient-final-solution-renders}b compared to row \ref{fig:patient-intermediate-solution-renders}b can be observed.
However, the overall changes in contour quality and deformations are minimal, indicating that the results obtained after 5~minutes are largely indicative of final achieved results.

\subsection{Quantitative comparison}
In Table~\ref{tab:approach-table}, we list the distance errors of the transformed bladder contours and the inverted transformation volumes.
We observe that the contour distances are often larger for DL-MODIR than for \mbox{(DL-)MOREA} or comparable for DL-MODIR and \mbox{(DL-)MOREA}.
On all patients, DL-MODIR requires larger folded volumes than \mbox{(DL-)MOREA} to achieve its transformations, with the small (0--8~\textit{ml}) folded volumes of \mbox{(DL-)MOREA} likely attributable to the low image resolution compared to the granularity of the mesh.
DL-MOREA yields smaller median contour distances after 5~minutes than MOREA, or equal median contour distances after 5~minutes, for all tested patients, but according to the Mann-Whitney U tests listed in Table~\ref{tab:stat-test-table}, this improvement is only significant in 4~patients, all belonging to the group of 6~patients with a large relative change in bladder volume.
After convergence, this difference is significant only in 2~patients.

In Figure~\ref{fig:convergence}, we show convergence plots of the best found organ contour distance objective value per generation across all patients and MOREA and DL-MOREA repeats.
These convergence plots generally confirm the visual observation that the strongest improvements in contour distance are made at the beginning of the optimization, with diminishing returns as optimization progresses.
It should be noted that sole inspection of this convergence trajectory gives an incomplete picture of the overall convergence, as other objectives and the overall spread of the approximation front are neglected, and should be considered jointly with the preceding quantitative and qualitative analyses.
When considering the optimization runtime required to reach convergence, we find that the original MOREA approach requires runtimes with an interquartile range (IQR) of 40 to 49 minutes to converge on patients in this dataset.
For DL-MOREA, runtimes have an IQR of 36 to 44 minutes, as DL-MOREA typically requires fewer mesh repair operations during optimization.

\begin{landscape}
\begin{table}[t]
    \footnotesize
    \centering
    \setlength{\tabcolsep}{4pt}
    \rowcolors{2}{gray!15}{white}

    \caption{\footnotesize Contour distances and inverted volumes of all three approaches, for automatically selected transformations. For each patient, the relative bladder volume change cluster is indicated. The median and interquartile range is reported for all configurations that are repeated with different random seeds. For MOREA and DL-MOREA, results are reported after 5 min. and after full convergence (\textit{final}). \textit{Legend:} D = Hausdorff 95th percentile bladder contour distance; I = inverted volume in the inverse DVF.}
    \label{tab:approach-table}
    
    \begin{tabular}{l|rr|rrrr|rrrr}
    \rowcolor{white}
    \toprule
Patient no.  & \multicolumn{2}{c|}{\textbf{DL-MODIR}} & \multicolumn{4}{c|}{\textbf{MOREA}} & \multicolumn{4}{c}{\textbf{DL-MOREA}} \\
{\scriptsize (relative} & & & \multicolumn{2}{c}{\textbf{After 5 min.}} & \multicolumn{2}{c|}{\textbf{Final (40-49 min.)}} & \multicolumn{2}{c}{\textbf{After 5 min.}} & \multicolumn{2}{c}{\textbf{Final (36-44 min.)}} \\
    \rowcolor{white}
{\scriptsize volume change)} & D [mm] & I [ml] & D [mm] & I [ml] & D [mm] & I [ml] & D [mm] & I [ml] & D [mm] & I [ml] \\
\midrule
62 (small) &   3.0 &  47.8 &   5.2 [4.2, 5.2] &   0.4 [0.4, 0.5] &   4.2 [4.2, 4.2] &   0.3 [0.2, 0.3] &   5.2 [5.2, 5.2] &   0.2 [0.2, 0.4] &   4.2 [4.2, 4.2] &   0.5 [0.3, 0.5] \\
63 (small) &   4.2 &  10.2 &   6.0 [5.2, 6.0] &   0.0 [0.0, 0.0] &   6.0 [6.0, 6.0] &   0.0 [0.0, 0.0] &   5.2 [5.2, 6.0] &   0.0 [0.0, 0.1] &   5.2 [5.2, 6.0] &   0.0 [0.0, 0.0] \\
64 (large) &  40.5 & 163.8 &  20.3 [19.0, 21.2] &   7.0 [5.2, 7.4] &   9.5 [9.5, 12.7] &   8.1 [7.7, 8.3] &  17.0 [15.6, 21.2] &   5.9 [5.8, 8.1] &   9.0 [9.0, 9.5] &   7.9 [6.9, 8.2] \\
65 (large) &  39.0 & 224.9 &  27.7 [27.0, 28.3] &   4.3 [4.0, 4.4] &  12.4 [10.8, 12.4] &   5.6 [4.3, 5.6] &  24.9 [22.0, 26.8] &   5.8 [3.7, 8.1] &   9.9 [9.5, 12.0] &   8.0 [4.0, 8.7] \\
66 (large) &  33.7 & 150.8 &  18.5 [18.5, 21.0] &   2.7 [1.2, 3.3] &   7.3 [7.3, 9.0] &   1.2 [0.9, 1.3] &  12.4 [12.4, 13.4] &   1.8 [1.8, 1.9] &   6.7 [6.0, 6.7] &   1.5 [1.2, 1.6] \\
67 (small) &   6.7 &   7.7 &   5.2 [4.2, 5.2] &   0.0 [0.0, 0.1] &   4.2 [4.2, 4.2] &   0.0 [0.0, 0.0] &   4.2 [4.2, 4.2] &   0.0 [0.0, 0.0] &   4.2 [4.2, 4.2] &   0.0 [0.0, 0.0] \\
68 (medium) &  18.2 &  21.9 &   4.2 [4.2, 5.2] &   0.6 [0.2, 1.3] &   4.2 [4.2, 4.2] &   0.4 [0.4, 0.5] &   4.2 [4.2, 4.2] &   1.1 [0.4, 1.2] &   4.2 [4.2, 4.2] &   0.2 [0.1, 0.5] \\
69 (medium) &  12.0 &  64.9 &   7.3 [6.7, 7.3] &   0.1 [0.1, 0.2] &   6.7 [6.7, 6.7] &   0.1 [0.1, 0.2] &   7.3 [6.7, 7.3] &   0.2 [0.2, 0.6] &   6.7 [6.7, 6.7] &   0.2 [0.1, 0.2] \\
70 (small) &   6.0 &  34.5 &   6.7 [6.7, 7.3] &   1.2 [0.5, 1.8] &   6.7 [6.7, 6.7] &   0.6 [0.5, 1.4] &   6.7 [6.0, 6.7] &   0.5 [0.2, 0.8] &   6.7 [6.7, 6.7] &   0.4 [0.1, 0.5] \\
71 (small) &   3.0 &   9.6 &   5.2 [4.2, 5.2] &   0.0 [0.0, 0.0] &   5.2 [4.2, 5.2] &   0.0 [0.0, 0.1] &   4.2 [4.2, 5.2] &   0.0 [0.0, 0.0] &   4.2 [4.2, 5.2] &   0.1 [0.0, 0.1] \\
72 (large) &  54.0 & 251.1 &  39.0 [38.2, 44.3] &   3.2 [2.3, 5.3] &  13.4 [12.4, 15.3] &   2.9 [2.7, 6.6] &  32.0 [31.9, 32.3] &   4.6 [3.0, 5.6] &  10.8 [10.8, 12.0] &   3.9 [3.8, 4.0] \\
73 (small) &  10.4 &  72.6 &   6.0 [5.2, 6.0] &   0.1 [0.0, 0.5] &   5.2 [5.2, 5.2] &   0.1 [0.1, 0.1] &   5.2 [5.2, 6.0] &   0.0 [0.0, 0.0] &   4.2 [4.2, 4.2] &   0.1 [0.1, 0.2] \\
74 (large) &  40.9 & 226.0 &  24.0 [24.0, 27.0] &   2.4 [2.4, 3.7] &  15.0 [14.7, 15.0] &   3.1 [2.2, 3.3] &  15.0 [15.0, 16.2] &   5.8 [4.1, 5.8] &  12.0 [12.0, 12.7] &   1.7 [1.4, 3.2] \\
75 (large) &  27.7 & 145.6 &  15.6 [15.0, 16.2] &   2.3 [0.8, 2.6] &   9.0 [8.5, 9.5] &   1.5 [1.1, 1.7] &  12.7 [12.4, 17.5] &   3.1 [2.8, 3.6] &   9.0 [8.5, 9.0] &   0.5 [0.5, 0.7] \\
76 (medium) &   9.0 &  47.3 &   6.7 [6.0, 6.7] &   2.2 [1.9, 2.8] &   6.0 [6.0, 6.0] &   0.8 [0.8, 1.0] &   6.0 [6.0, 6.0] &   0.9 [0.9, 1.3] &   6.0 [6.0, 6.0] &   0.5 [0.4, 1.1] \\
\bottomrule
    \end{tabular}
\end{table}
\end{landscape}

\begin{table}[t]
    \centering
    \caption{Comparison of Hausdorff 95th percentile bladder contour distances, between MOREA and the proposed hybrid, DL-MOREA. The relative change between the medians of two distance distributions for a certain patient and time point is reported (i.e., $1-(H_{\text{MOREA}} / H_{\text{DL-MOREA}})$, where $H_x$ is the median distance for a certain approach configuration $x$). Between parentheses, the results of a Mann-Whitney U test are given, which indicate the difference of distributions is significant. Significant results (according to an alpha level of 0.05) are highlighted in bold.}
    \label{tab:stat-test-table}
    \begin{tabularx}{0.75\textwidth}{Xrr}
\toprule
& \multicolumn{2}{c}{Relative difference of contour distance} \\
Patient no. (relative volume change) & After 5 min. & Final \\
\midrule
Patient 62 (small) & {    0\%} ($p=0.734$) & {    0\%} ($p=0.424$) \\
Patient 63 (small) & {   13\%} ($p=0.631$) & {   13\%} ($p=0.270$) \\
Patient 64 (large) & {   17\%} ($p=0.346$) & {    5\%} ($p=0.125$) \\
Patient 65 (large) & \textbf{   10\%} ($p=0.036$) & {   20\%} ($p=0.292$) \\
Patient 66 (large) & \textbf{   33\%} ($p=0.011$) & \textbf{    9\%} ($p=0.023$) \\
Patient 67 (small) & {   18\%} ($p=0.270$) & {    0\%} ($p=0.424$) \\
Patient 68 (medium) & {    0\%} ($p=0.177$) & {    0\%} ($p=0.424$) \\
Patient 69 (medium) & {    0\%} ($p=0.817$) & {    0\%} ($p=1.000$) \\
Patient 70 (small) & {    0\%} ($p=0.076$) & {    0\%} ($p=0.424$) \\
Patient 71 (small) & {   18\%} ($p=0.734$) & {   18\%} ($p=0.631$) \\
Patient 72 (large) & \textbf{   18\%} ($p=0.016$) & {   19\%} ($p=0.115$) \\
Patient 73 (small) & {   13\%} ($p=0.373$) & \textbf{   18\%} ($p=0.023$) \\
Patient 74 (large) & \textbf{   38\%} ($p=0.012$) & {   20\%} ($p=0.089$) \\
Patient 75 (large) & {   18\%} ($p=0.841$) & {    0\%} ($p=1.000$) \\
Patient 76 (medium) & {   11\%} ($p=0.067$) & {    0\%} ($p=0.424$) \\
\bottomrule
    \end{tabularx}
\end{table}

\section{DISCUSSION AND CONCLUSIONS}

For the first time, a DL-based DIR approach has been combined with a biomechanical DIR approach.
In addition, this work is also the first to achieve this with a multi-objective approach.
The proposed hybrid, called DL-MOREA, is initialized with the predicted transformations of an existing DL DIR model and multi-objectively optimizes these with a biomechanical mesh transformation model.
On pelvic CT scans with large deformations, experiments show that the proposed hybrid is able to improve on the folding and bladder contour distance performance of the original DL-based DIR approach, within 5~minutes.
The added value of the hybrid is greatest in difficult registration cases, although it also provides benefits in easier cases, e.g., through folding reduction.
DL-MOREA therefore shows promise to combine the best of both worlds in a multi-objective setting: the speed of DL-based registration and the realism of biomechanical registration.

There are limitations to the validity of this experimental study.
For example, the process of selecting a registration from each approximation set for comparison obscures the complexity of a true multi-objective comparison of approximation sets.
Furthermore, the qualitative comparison of registrations was not performed by a group of multiple experts, leading to potential observer bias.
A third limitation of this comparison is the used dataset, which has a limited image resolution and field of view.
However, image features and structures that are important to the region of interest remain visible in the dataset.
Finally, the limited size of the training dataset poses another potential limitation.

Future work should address these concerns by involving multiple expert observers in an \textit{a posteriori} manual selection and assessment process, to model the intended real-world use more closely.
The full approximation sets should also be compared using independent metrics, to evaluate their diversity and quality in a quantitative comparison.
In general, the proposed hybrid should also be evaluated across other registration tasks and anatomical sites to assess how its performance generalizes.
Furthermore, hybrid combinations of other DIR approaches should also be studied, to investigate the impact of the DIR components chosen in this work.

\acknowledgments
 
The authors thank W.~Visser-Groot and S.M.~de Boer MD, PhD (Dept. of Radiation Oncology, Leiden University Medical Center, Leiden, The Netherlands) for their contributions to this study.
The authors also thank M.~Grewal (Evolutionary Intelligence Group, Centrum Wiskunde \& Informatica, Amsterdam, The Netherlands) for her guidance on the application of the DL-MODIR approach.
This research is part of the research programme Open Technology Programme with project number 15586, which is financed by the Dutch Research Council (NWO), Elekta, and Xomnia. 
Further, this work is co-funded by the public-private partnership allowance for top consortia for knowledge and innovation (TKIs) from the Ministry of Economic Affairs.


\bibliography{references} 

\begin{thebibliography}{10}

\bibitem{Pirpinia2017}
Pirpinia, K., Bosman, P. A.~N., Loo, C.~E., Winter-Warnars, G., Janssen, N. N.~Y., Scholten, A.~N., Sonke, J.~J., van Herk, M., and Alderliesten, T., ``{The feasibility of manual parameter tuning for deformable breast MR image registration from a multi-objective optimization perspective},'' {\em Physics in Medicine and Biology}~{\bf 62}(14),  5723--5743 (2017).

\bibitem{Eppenhof2020}
Eppenhof, K. A.~J., Maspero, M., Savenije, M. H.~F., de~Boer, J.~C., van der Voort~van Zyp, J. R.~N., Raaymakers, B.~W., Raaijmakers, A. J.~E., Veta, M., van~den Berg, C. A.~T., and Pluim, J. P.~W., ``{Fast contour propagation for MR-guided prostate radiotherapy using convolutional neural networks},'' {\em Medical Physics}~{\bf 47},  1238--1248 (3 2020).

\bibitem{Salehi2022}
Salehi, M., Vafaei~Sadr, A., Mahdavi, S.~R., Arabi, H., Shiri, I., and Reiazi, R., ``{Deep Learning-based Non-rigid Image Registration for High-dose Rate Brachytherapy in Inter-fraction Cervical Cancer},'' {\em Journal of Digital Imaging} ,  1--14 (2022).

\bibitem{DeVos2019}
de~Vos, B.~D., Berendsen, F.~F., Viergever, M.~A., Sokooti, H., Staring, M., and I{\v{s}}gum, I., ``{A deep learning framework for unsupervised affine and deformable image registration},'' {\em Medical Image Analysis}~{\bf 52},  128--143 (2019).

\bibitem{Grewal2024}
Grewal, M., Westerveld, H., Bosman, P. A.~N., and Alderliesten, T., ``{Multi-Objective Learning for Deformable Image Registration},'' in [{\em Medical Imaging with Deep Learning}{\nolinebreak\hspace{0.1em}]},   {\bf 178},  1--16 (2024).

\bibitem{Brock2005}
Brock, K.~K., Sharpe, M.~B., Dawson, L.~A., Kim, S.~M., and Jaffray, D.~A., ``{Accuracy of finite element model-based multi-organ deformable image registration},'' {\em Medical Physics}~{\bf 32}(6),  1647--1659 (2005).

\bibitem{Zhong2012}
Zhong, H., Kim, J., Li, H., Nurushev, T., Movsas, B., and Chetty, I.~J., ``{A finite element method to correct deformable image registration errors in low-contrast regions},'' {\em Physics in Medicine and Biology}~{\bf 57}(11),  3499--3515 (2012).

\bibitem{Zhang2004}
Zhang, T., Orton, N.~P., Mackie, T.~R., and Paliwal, B.~R., ``{Technical note: A novel boundary condition using contact elements for finite element deformable image registration},'' {\em Medical Physics}~{\bf 31}(9),  2412--2415 (2004).

\bibitem{Andreadis2023}
Andreadis, G., Bosman, P. A.~N., and Alderliesten, T., ``{MOREA: a GPU-accelerated Evolutionary Algorithm for Multi-Objective Deformable Registration of 3D Medical Images},'' in [{\em Proceedings of the 2023 Genetic and Evolutionary Computation Conference}{\nolinebreak\hspace{0.1em}]},   1294--1302 (2023).

\bibitem{Deist2023}
Deist, T.~M., Grewal, M., Dankers, F. J. W.~M., Alderliesten, T., and Bosman, P. A.~N., ``{Multi-objective Learning Using HV Maximization},'' in [{\em Proceedings of the 12th International Conference on Evolutionary Multi-Criterion Optimization (EMO)}{\nolinebreak\hspace{0.1em}]},   103--117 (2023).

\bibitem{Balakrishnan2019}
Balakrishnan, G., Zhao, A., Sabuncu, M.~R., Guttag, J., and Dalca, A.~V., ``{VoxelMorph: A Learning Framework for Deformable Medical Image Registration},'' {\em IEEE Transactions on Medical Imaging}~{\bf 38}(8),  1788--1800 (2019).

\bibitem{Zitzler2008}
Zitzler, E., Knowles, J., and Thiele, L., ``{Quality Assessment of Pareto Set Approximations},'' in [{\em Multiobjective Optimization}{\nolinebreak\hspace{0.1em}]},   373--404, Springer Berlin Heidelberg (2008).

\bibitem{Andreadis2022}
Andreadis, G., Bosman, P. A.~N., and Alderliesten, T., ``{Multi-objective dual simplex-mesh based deformable image registration for 3D medical images - proof of concept},'' in [{\em SPIE Medical Imaging 2022: Image Processing}{\nolinebreak\hspace{0.1em}]},   744--750 (2022).

\end{thebibliography}
\bibliographystyle{spiebib} 

\end{document}